\documentclass[10pt, a4paper]{article}
\usepackage{lrec2022} 
\usepackage{multibib}
\newcites{languageresource}{Language Resources}
\usepackage{graphicx}
\usepackage{tabularx}
\usepackage{soul}

\usepackage{titlesec}
\titleformat{\section}{\normalfont\large\bfseries\center}{\thesection.}{1em}{}
\titleformat{\subsection}{\normalfont\SmallTitleFont\bfseries\raggedright}{\thesubsection.}{1em}{}
\titleformat{\subsubsection}{\normalfont\normalsize\bfseries\raggedright}{\thesubsubsection.}{1em}{}
\renewcommand\thesection{\arabic{section}}
\renewcommand\thesubsection{\thesection.\arabic{subsection}}
\renewcommand\thesubsubsection{\thesubsection.\arabic{subsubsection}}

\usepackage{epstopdf}
\usepackage[utf8]{inputenc}

\usepackage{hyperref}
\usepackage{xstring}

\usepackage{color}
\newcommand{\quotes}[1]{``#1''}

\title{A Benchmark Corpus for the Detection of Automatically Generated Text in Academic Publications}

\name{Vijini Liyanage, Davide Buscaldi, Adeline Nazarenko} 

\address{Université Sorbonne Paris Nord, LIPN, CNRS, UMR 7030  \\
         F‐93430, Villetaneuse, France \\
         \{liyanage, davide.buscaldi, adeline.nazarenko\}@lipn.univ-paris13.fr\\}

\abstract{
Automatic text generation based on neural language models has achieved performance levels that make the generated text almost indistinguishable from those written by humans. Despite the value that text generation can have in various applications, it can also be employed for malicious tasks. The diffusion of such practices represent a threat to the quality of academic publishing. To address these problems, we propose in this paper two datasets comprised of artificially generated research content: a completely synthetic dataset and a partial text substitution  dataset. In the first case, the content is completely generated by the GPT-2 model after a short prompt extracted from original papers. The partial or hybrid dataset is created by replacing several sentences of abstracts with sentences that are generated by the Arxiv-NLP model. We evaluate the quality of the datasets comparing the generated texts to aligned original texts using fluency metrics such as BLEU and ROUGE. The more natural the artificial texts seem, the more difficult they are to detect and the better is the benchmark. We also evaluate the difficulty of the task of distinguishing original from generated text by using state-of-the-art classification models. 
 \\ \newline  }

\begin{document}

\maketitleabstract

\section{Introduction}

The Transformer model \cite{Vaswani-Shazeer-Parmar-Uszkoreit-Jones-Gomez-Kaiser-Polosukhin-17} can be considered as a major milestone in the domain of deep learning and Natural Language Processing (NLP). Since then, various forms of state-of-the-art transformer models such as the Generative Pre-training model (GPT) \cite{Radford-Narasimhan-Salimans-Sutskever-18}, BERT \cite{Devlin-Chang-Lee-Toutanova-18} and Transformer-XL \cite{Dai-Yang-Yang-Carbonell-Le-Salakhutdinov-19} have been introduced and utilized for a diverse amount of NLP tasks. To cite some, Natural Language Generation (NLG) \cite{radford2019language}, text classification \cite{yang2019xlnet}, machine translation \cite{lample2019cross} and text summarization \cite{lewis2019bart}. The aforementioned research show that the transformer models are capable of producing outstanding results. \par


The GPT model has been recently updated to significantly improve the quality of the generated text. For example, the GPT-2 model \cite{radford2019language} is competent enough to produce natural text that looks like as if written by a human \cite{solaiman2019release}. Regardless of the valuable contribution provided by these models for the betterment of NLP in general, some concerns have been raised about the a potential risks associated to such models. These models can be misused for malicious tasks such as fake news generation \cite{zellers2019defending}, \cite{vosoughi2018spread}, fake review generation \cite{adelani2020generating} and viral story generation \cite{faris2017partisanship}, \cite{wardle2017information}. 

Recently, some of these models have been applied for the computer-assisted writing of research papers \cite{wang2019paperrobot}, reviews \cite{wang2020reviewrobot} or theses \cite{abd2019artificial}. Despite the advantages in alleviating researchers' workload, a risk for misuse of these technologies exists. A recent work \cite{cabanac2021prevalence} shows  that old textual generation models based on context-free grammars, such as SciGen\footnote{\url{https://github.com/strib/scigen.git}}, are being actively used in academic publishing, although their detection is relatively easy since they tend to produce nonsense or ``tortured phrases" -- weirdly paraphrased versions of scientific terms. Therefore, immediate research on detection of academic texts that are artificially generated is imperative. In this way, researchers will also have a tool to determine whether these powerful models were used in a responsible way or not. To develop such detection methods, there is a vital requirement of a corpus composed of automatically generated academic content. This is the main purpose of the present research.

\section{Related Work}

\subsection{Human Detection of Machine Generated Text}

Several research works have been conducted regarding human-detection of machine generated text. 
\cite{bakhtin2019real} considers human-detection as a ranking task. Based on 
a thorough analysis of how human detection is affected by factors such as sampling method and the length of the text excerpt, the authors of \cite{ippolito2019human} consider that human detectors perform well in finding semantic errors in machine generated text. \cite{gehrmann2019gltr} claims that the accuracy of human detection of artificially generated text without any tool is only 54\%. 
\cite{ippolito2019automatic} has identified that humans focus mostly on semantic errors, while the discriminative models such as fine-tuned BERT focus more on statistical artifacts. One of the latest tools, RoFT (Real or Fake Text) tool \cite{dugan2020roft}, is used to evaluate human detection showing 
that the text generation models are capable to fool humans by one or two sentences. A recent research \cite{clark2021all} shows that 
training humans on evaluation task for GPT-3 generated text only improves the overall accuracy upto 50\%.
Despite the interest in measuring the ability of humans to detect automatically generated text, not much research has been conducted to develop automatic tools to distinguish machine generated text from human written text. 

\subsection{Automatic Detection of Machine Generated Text}

The introduction of the statistical model GLTR (Giant Language model Test Room) \cite{gehrmann2019gltr} can be considered as a major milestone for the detection of automatically generated text. The authors consider the stylometric details of texts by incorporating three types of tests: the probability of the word, the absolute rank of a word and the entropy of the predicted distribution. Afterwards, they compute per-token likelihoods and visualize histograms over them to support humans in detection of automatically generated content. A recent research \cite{al2020identification} has extended the work done in GLTR by feeding the output of GLTR to a Convolutional Neural Network, which automatically classifies whether the input reviews are human written or machine generated. \par

Another turning point is the establishment of the GROVER \cite{zellers2019defending}, in which it's architecture is a combination of a generation model and detection model. News are generated 
using a transformer based model which has an architecture similar to GPT-2 \cite{radford2019language}. But \cite{radford2019language} have used conditional generation (on article meta data) and nucleus sampling. Afterwards, a zero shot detection is performed using a simple linear classifier on top of the pre-trained GROVER model. The authors have experimented detection with existing other models as well( fastText \cite{bojanowski2017enriching} and BERT\cite{Devlin-Chang-Lee-Toutanova-18}) reporting the highest accuracy for their own GROVER model and claiming that the best models in forming fake content are also the best models in detection.  
Contrary to that, Uchendu shows, however, that GROVER cannot correctly detect texts generated by language models other than GROVER itself. 

Lots of research have leveraged the RoBERTa \cite{liu2019roberta}, a masked and non-generative language model to detect automatically generated text. \cite{solaiman2019release} has proved that the discriminative model of RoBERTa 
outperforms generative models such as GPT-2 in detection task. Such 
findings contradict with 
GROVER authors' claim 
that the generative model is better in detecting text generated by itself. \cite{fagni2021tweepfake} reveals that the RoBERTa can defeat traditional machine learning models, such as bag of words, and neural network models, such as RNNs and CNNs, regarding the detection of automatically generated tweets. Moreover, \cite{uchendu2020authorship} shows that RoBERTa outperforms existing detectors in detecting automatically generated news articles and product reviews which are generated by state of the art models like GPT-2. Despite the success of RoBERTa, recent research \cite{jawahar2020automatic} shows that its dependence on large amounts of data, limits limits its use for detection. 
\cite{wolff2020attacking} challenges the RoBERTa model by exposing it with homoglyph and misspelling attacks and their results show a drastic drop in recall.  
This shows that the future detection models should be robust against such attacks. To test and evaluate the detection models, it is of key importance to have a dataset that incorporates such attacks or traps, if possible independent of any specific detection method.


Many of the research on detection assumes that the generator is known (\cite{gehrmann2019gltr}, \cite{zellers2019defending}). Therefore their approaches are incapable in generalizing the settings so that it works well when the generator architectures and corpora are different in training and testing stages. However, in real world settings, a detection model faces indeterminate and evolving data. This issue has been addressed to a certain level by \cite{bakhtin2019real}, which provides a thorough experimental setup and quantitative results. \cite{ippolito2019automatic}  fine-tunes the BERT model for detection and analyzes how factors like sampling strategies and text excerpt length impact the detection task. Another research \cite{varshney2020limits} produces a formal hypothesis testing framework and sets error exponents limits (in terms of perplexity and cross-entropy) for large scale models such as  GPT-2 and CTRL in order to find limits in detecting text generated by them. One of the latest research \cite{maronikolakis2020identifying} leverages Transformers to detect headlines generated by GPT-2 model. 
In this approach, the models learn from the datasets and classifies text as machine generated text or human-written text. It makes use of 
4 types of classifiers:
Baselines (Logistic Regression, Elastic Net), Deep learning (CNN, Bi-LSTM, Bi-LSTM with Attention), Transfer learning via ULMFit and Transformers (BERT, DistilBERT) for the detection task. The results show that BERT scores an overall accuracy of 85.7\%.

In addition to transformer based models many research works conducted for the detection make use of other types of deep learning models as well as statistical models. \cite{vijayaraghavan2020fake} experiments numeric representation such as Tf-idf and word2vec, as well as neural networks such as ANNs and LSTMs on detection of fake news. A latest research \cite{harada2021discrimination}  suggests two approaches -- CPCO (Consistency of anti preceding sentence using Cosine words Overlapping),
 and CICO (Consistency of opposing Input sentences using Cosine words Overlapping) --, which utilize sentence coherence for the detection task. Also \cite{jawahardetecting}  leverages different discourse models for detection. By exposing style-based classifiers to syntactic and semantic permutations, \cite{bhat2020effectively} shows the limitations of style-based classifiers which highly rely on provenance to detect fake text. Furthermore, \cite{perez2017automatic} highlights the importance of linguistic features such as semantic, syntactic and lexical features in distinguishing the machine generated news from human written news.

 Although there are lots of research conducted for the detection of automatically generated content such as news paper articles, reviews, tweets, headlines and so on, only a  limited number of detection research is dedicated to academia. These works \cite{xiong2009effective,lavoie2010algorithmic} have mainly focused on the authenticity of the references. \cite{amancio2015comparing} has examined topological properties in natural and generated papers as a source for detection.   \cite{williams2015use} has proposed various measures in detecting generated academic content. SciDetect \cite{nguyen2016engineering} is another research which considers inter-textual distance using all the words and nearest neighbour classification for detection. The authors have analyzed the existing methods for detecting automatically generated paper and relied on  Probabilistic Context Free Grammar (PCFG) to detect fake academia. However, their approach relies on the fact that word distributions are identical to that of human written content, an assumption that is not always verified. 
 A recent research \cite{cabanac2021prevalence} proposes a rule based detection mechanism which leverage third party search engines to distinguish automatically generated papers based on improbable word sequences found in them, but their approach can only detect grammar based computer generated papers. However, among all the works about the detection of automatically generated text, there is no research on texts in the academic or scholarly domain, despite the availability of such data and the potential danger in the misuse of generative models in this domain.
 
 In order to leverage the deep learning models to detect automatically generated research content from human written content, a corpus of artificially generated academic content is required. Many latest research \cite{cabanac2021prevalence}, \cite{xiong2009effective} have leveraged SCIgen to generate a dataset for their research. But SCIgen generates nonsense (because it focuses 
 on amusement rather than coherence) using context free grammar. To test methods and encourage research in this area, we propose a benchmark of academic papers that are generated using state-of-the-art models like GPT-2 an that look as if they were written by humans. This paper presents the dataset that composes this benchmark, explain how it has been built and evaluate its usefulness for the detection of automatically generated academic texts. 

\section{Benchmark datasets}
To the best of our knowledge, there are no publicly available corpora which can be used as a benchmark dataset to experiment the detection of automatically generated academic content. However, as explained above, such a benchmark is a prerequisite for any research in the field of fake text detection, which represents an important issue for the academic world. This paper presents datasets for this purpose.

We built two corpora: one containing automatically generated papers and 
a hybrid dataset which contains original (human written) abstracts in which some sentences are substituted with machine generated sentences. We designed them as a mixture of natural (written by humans) and artificial (generated) content that should be not easily detected by a machine and we used different generation strategies. In order to achieve this, we had to tune the models to generate content in a window of “credibility”, or in other words that the generated content could appear as original to an uninformed reader. The automatically generated paper dataset and the hybrid dataset contain average word counts of 1243 and 177, respectively.


The first corpus is composed of artificially generated research papers, aligned to the original abstract from which the writing prompt was chosen. It corresponds to the situation in which a malicious author would create a full text to submit, in a similar way to Sci-Gen. Their length is variable as they can be composed from the abstract alone to more sections such as introduction, related work and conclusion. We eliminated sections such as Methodology and Results and Evaluation, because they mostly contain diagrams and tables which can reveal too easily their (synthetic) origin. Few examples showing content from the original paper and the respective generated paper are available in Table \ref{table1}. Here the original papers correspond to the papers which were used as prompts for the models to perform generation.

\begin{table*}[ht]
 \begin{center}
 \begin{tabularx}{\textwidth}{|X|X|}

      \hline
      Original Paper&Generated Paper\\
      \hline
      Finally, in this task, we used comparative evaluations to \textbf{determine} the quality of the sets of generated \textbf{KPs. Obviously, this results with rankingbased evaluation. Establishing an absolute evaluation metric in this context merits further investigation. We thank all participating teams for taking the time to participate in this challenging new shared task. We believe that these contributions, along with the data collected and shared in this report, will be valuable to further advance the research around KPA and related tasks.} & Finally, in this task, we used comparative evaluations to \textbf{estimate} the quality of the sets of generated \textbf{texts. We expect the task and the findings reported in this paper to be relevant for researchers working on text summarization and argument mining. Introduction Key Point Analysis (KPA) was introduced in BarHaim et al. (2020a,b) as a challenging NLP task with tight relations to Computational Argumentation, Opinion Analysis, and Summarization, and with many practical applications (Bar-Haim et al., 2021).}\\
      \hline
       We show that the mixed-domain transfer learning approach \textbf{does not always work well. We found} the \textbf{combination} of the \textbf{non-graph DSM (for capturing local relationships) and CNN (for extracting global relationships) to be a promising approach towards creating a gener- alizable solution for detecting COVID-19 online misinformation. In the future, we plan to investigate the gener- alizability of the DSM models created using deep learning architectures such as BERT.}
       &
       We show that the mixed-domain transfer learning approach \textbf{is more effective across most of} the \textbf{dimensions} of the \textbf{COVID-19 data. Specifically, we show that the shallow mixed-domain transfer learning approach is more effective across the following dimensions of the COVID-19 data: granular (grained), shallow (skewed), and deep (deeper). We show that the shallow mixed-domain transfer learning approach is more effective across the following dimensions of the COVID-19 data: temporal dimension (the context in the dataset evolves), length dimension.}
       \\
     
      \hline

\end{tabularx}
\caption{Some examples of original \emph{vs.} generated papers in the ``fully generated" corpus.}
\label{table1}
 \end{center}
\end{table*}

The second corpus is a collection of abstracts which are composed of a combination of original content and machine generated sentences. It corresponds to a situation in which an author would recur to language generation to fill in certain parts of hers paper. To compose this corpus, we ignored papers containing the name of the proposed model, product or project in the proposal statement of the abstract, as the Arxiv-NLP model might suggest a name that would not be consistent with the rest of the abstract (the sentences extracted from the original abstract), thus making the generated abstract too easily distinguishable from human written abstracts. Table~\ref{table2} shows some examples of the original abstracts and their corresponding generated abstracts.

\begin{table*}[ht]
 \begin{center}
 \begin{tabularx}{\textwidth}{|X|X|}

      \hline
      Original Abstract&Generated Abstract\\
      \hline
      Our experiments suggest \textbf{that models possess belief-like qualities to only a limited extent, but update methods can both fix incorrect model beliefs and greatly improve their consistency.} Although off-the- shelf optimizers are surprisingly strong belief- updating baselines, our learned optimizers can outperform them in more difficult settings than have been considered in past work. 
      & 
      Our experiments suggest \textbf{the importance of model beliefs in  learning models, and we show that the approach outperforms  automatic model updating systems using word representations.} Although off-the- shelf optimizers are surprisingly strong belief- updating baselines, our learned optimizers can outperform them in more difficult settings than have been considered in past work.\\
      \hline
      Simultaneously evolving morphologies (bodies) and controllers (brains) of robots can cause a mismatch between the inherited body and brain in the offspring. To mitigate this problem, the addition of an infant learning period by the so-called Triangle of Life framework has been proposed relatively long ago. However, an empirical assessment is still lacking to-date. In this paper we \textbf{investigate the effects of such a learning mechanism from different perspectives.} 
      & 
      Simultaneously evolving morphologies (bodies) and controllers (brains) of robots can cause a mismatch between the inherited body and brain in the offspring. To mitigate this problem, the addition of an infant learning period by the so-called Triangle of Life framework has been proposed relatively long ago. However, an empirical assessment is still lacking to-date. In this paper , \textbf{we present a method to evaluate the effect of an algorithm based on the development of a hybrid human/bot learning framework, which combines the development of both a hybrid robot and a human model on the same domain.} \\
      \hline
\end{tabularx}
\caption{Some examples of original \emph{vs.} generated abstracts from the ``hybrid" corpus.}
\label{table2}
 \end{center}
\end{table*}

The dataset is available at the following URL: \url{https://github.com/vijini/GeneratedTextDetection.git}.

\section{Benchmark Design Methodology}

For the first corpus (``fully generated"), we used the GPT-2 model for the generation of research papers and the model was fine-tuned by feeding it with papers extracted from ArXiv\footnote{\url{https://arxiv.org}} . We specifically selected the Computation and Language domain and the 100 latest papers were chosen to make sure that the papers we considered were not used to train the original GPT-2 model. Moreover, papers with IEEE citation style were not selected to train the model, because this style represent citations only as numbers, which does not allow a reader to verify whether the citations are appropriate (we do not consider the ``References" section for model training). Each paper was separately used to fine-tune the GPT-2 model. By choosing the first 50 words of each original paper as the seed text, a new paper which is of same length as the original paper was generated using the fine-tuned model. Likewise, 100 new papers were generated. The temperature parameter was set to 0.7 to make sure that the generated content are neither too much random nor too much alike the original paper. For clarity purposes, we ignored the sections such as methodology, results and evaluation and discussion which contain diagrams, tables, equations.

The second (``hybrid") corpus was generated by combining the authentic content and the machine generated content. For this task we utilized some of the latest abstracts from Artificial Intelligence domain in ArXiv. Each hybrid abstract is made of 4 parts. The initial content is extracted from an original abstract up to the point where it reveals about the proposal (\emph{e.g.} \quotes{In this paper,}, \quotes{We propose}, \quotes{Here we}). The next sentence is generated until the first full-stop, using the Arxiv-NLP provided by the Huggingface team \cite{wolf-etal-2020-transformers}. Then, the rest of the original abstract is copied until the point that corresponds to the conclusion (\emph{e.g.} \quotes{We conclude}, \quotes{Our results show that} ). Again, using the Arxiv-NLP, the rest of the abstract is generated. Likewise 100 new abstracts were composed. The temperature parameter was set to 1 and top-p was 0.9. This generation is done with human intervention, so that it is biased towards the objective strategy of making the generated content difficult to detect. 


\section{Evaluation} \label{Results}

We present a double approach to evaluate the utility of the produced corpora for the task of classifying artificially generated and human written academic texts in a context where neural-based generation models have become common. We first evaluate the intrinsic quality of the generated texts, assuming that the more natural they seem the more difficult and the more misleading they are for the detection methods. We also perform an application based evaluation using a panel of state-of-art detection models to assess the difficulty of the classification task and to check that our benchmark is not biased towards a specific detection approach.


\subsection{Evaluation of the quality of the generated texts}

To assess the instrinsic quality of our benchmark, we 
leveraged two metrics: BLEU (Bilingual Evaluation Understudy) \cite{papineni2002bleu} and ROUGE (Recall-Oriented Understudy for Gisting Evaluation) \cite{lin2004rouge} to compare the generated contents with their respective original contents. BLEU and ROUGE are two traditional metrics used to compare the candidate (generated) text with the reference (original) text. They are traditionally considered as fluency metrics that indicates how natural an artificial text is compared to a natural original one. BLEU is a precision based score while ROUGE is a recall based score. We calculated the BLEU score 
 at uni-gram and sentence levels and ROUGE 
 at uni-gram (1), bi-gram (2) and Longest Common Sub-sequence (L) levels. All the results with respect to BLEU and ROUGE scores are provided in Table \ref{table3} and Table \ref{table4} for the fully generated  and hybrid datasets, respectively. For all the evaluations, 80/20 training/test split was done on the datasets.
 
 In general, all the average figures are quite high (more than 0.8), which indicates that the artificial texts are quite similar to natural ones. Of course come of the generated texts could be identified as ``copies" of the original texts but they have nevertheless been artificially generated and it is interesting to evaluate the detection models against these texts that ``look" natural. The tables represent both the minimum and average scores, so that a reader can get an idea about the quality of the overall dataset by the average score, while minimum score provides an idea of the quality of individual fully generated or hybrid research articles. Minimum \emph{vs.} average scores comparison shows that the methodology ensures a wide diversity of generated data: some artificial texts are quite different from the original ones and presumably easier to detect.
 
 Rouge-1 and Rouge-L scores are almost similar when they are rounded off to three decimal places. But when we consider more decimal places, it can be seen that the Rouge-1 scores are slightly higher than the Rouge-L scores. If we consider overall Unigram \emph{vs.} sentence BLEU, as the unigram BLEU metric is more tolerant than the sentence one, it is normal that the sentence BLEU scores, although quite high, are lower than the unigram BLEU scores.
 
 When we look at the results on Table \ref{table3}, it can be seen that the 
 uni-gram BLEU scores are higher than unigram ROUGE scores. On the other hand, when we measure the scores considering a longer sequence (either sentence level in BLEU or Longest Common Sub-sequence in ROUGE ), the BLEU scores are lower than ROUGE ones.
 
 When we look at the Table \ref{table4}, it can be seen that the average BLEU scores are always less than the respective ROUGE scores. When we refer the two tables \ref{table3} and \ref{table4}, it can be seen that the average BLEU scores are always higher in the fully generated dataset compared to hybrid dataset, which means the fully generated dataset has a better precision (that is, a good portion of the generated n-grams are also in the original text). On the other hand, it can be seen that the average ROUGE scores of the fully generated dataset is lower than the hybrid dataset. This means that the recall of the hybrid dataset is at a better level compared to the other dataset (this is expected as parts of the original text are included in the generated one). The minimum scores are relatively high for both BLEU and Rouge, indicating that there is a good degree of similarity between the texts even when they are most dissimilar. 

 \begin{table*}[ht]
 \begin{center}
 \begin{tabularx}{\textwidth}{|X|X|X|X|X|X|}
 
 \hline
 Score & unigram level BLEU & sentence BLEU & Rouge-1 & Rouge -2 & Rouge-L  \\
 \hline
 Min. Score & 0.659 & 0.583 & 0.591 & 0.559 & 0.591 \\
  \hline
 Avg. Score & 0.867 & 0.809 & 0.853 & 0.810 & 0.853 \\
 

   \hline
 
\end{tabularx}
\caption{Minimum and Average BLEU and ROUGE Scores (Recall) for the Fully Generated Dataset.}
\label{table3}
 \end{center}
\end{table*}
 
 \begin{table*}[ht]
 \begin{center}
 \begin{tabularx}{\textwidth}{|X|X|X|X|X|X|}
\hline
 Score & ngram level BLEU & sentence BLEU & Rouge-1 & Rouge -2 & Rouge-L  \\
 \hline
 Min. Score & 0.629 & 0.495 & 0.598 & 0.473 & 0.598 \\
  \hline
 Avg. Score & 0.824 & 0.792 & 0.882 & 0.840 & 0.881 \\ 
 \hline
\end{tabularx}
\caption{Minimum and Average BLEU and ROUGE Scores (Recall) for the Hybrid Dataset.}
\label{table4}
 \end{center}
\end{table*}

 

\subsection{Evaluation of the classification difficulty}

Since our datasets are to be used for benchmarking classification models, we also evaluate the difficulty of the classification task. 


To verify that the proposed benchmark is not biased towards a specific model type, we consider a variety of models, \emph{i.e.} 
statistical models such as Bag of Words \cite{harris1954distributional} as well as deep learning based models. As Bag of Words models, we consider Multinomial Naive Bayes, Passive Aggressive Classifier Multinomial Classifier with Hyper-parameter (alpha) algorithms and Support Vector Machine \cite{cortes1995support}. For the vocabulary, we considered not only single words but n-grams of size 1 - 3 words. As deep learning based models, we consider basic LSTM, Bi-LSTM (with the same configuration used by \cite{maronikolakis2020identifying}), BERT and DistilBERT \cite{sanh2019distilbert}.

The classification results are presented in Table ~\ref{table5} and Table~\ref{table6} for fully generated and hybrid datasets, respectively. As per the results, the highest classification score was obtained by the DistilBERT model regarding both the datasets: the scores are 62.5\% and 70.2\% for the fully generated dataset and the hybrid dataset, respectively. These results show that the generated corpora are competent enough to be used as a baseline datasets to experiment detection.


As expected, the scores differs from one model to the other, the deep learning based models having higher accuracy scores than the statistical models. The higher scores are obtained by DistilBERT model on both the datasets: 62.5\% and 70.2\% for the fully generated dataset and the hybrid dataset, respectively. Interestingly, if most models perform slightly better on the Hybrid Dataset, it is not the case of LSTM model which achieves a much better score on the Fully Generated Dataset and of BERT and passive aggressive classifier at lesser degrees. This is an argument for including both datasets and different types of generated data in our benchmark. Despite these differences, we observe globally that the accuracy scores are not very high, even for DistilBERT. This is the most important point to assess the quality of our benchmark in terms of classification difficulty.

We did a comparison of our results to  the latest research works \cite{maronikolakis2020identifying} and the results are depicted in Table \ref{table7}. The 
overall accuracies regarding our datasets are lower when compared with the aforementioned research. This 
may be due to the fact that Maronikolakis \emph{et al. focus on the generation of short content (headlines) but it shows that our datasets are more difficult to classify than theirs, which makes it a better benchmark proposal. }

 \begin{table*}[ht]
 \begin{center}
 \begin{tabularx}{\textwidth}{|l|X|}
\hline
 Model used & Accuracy \\
 \hline
 Bag of ngrams(1-3), Multinomial Naive Bayes Algorithm

 & 19.7 \\
  \hline
 Bag of ngrams(1-3),  Passive Aggressive Classifier Algorithm

 & 31.8 \\ 
  \hline
  Bag of ngrams(1-3), Multinomial Classifier with Hyperparameter (alpha) & 19.7\\
 \hline
 Bag of ngrams(1-3), SVM & 37.9\\
  \hline
  LSTM model & 59.1 \\
  \hline
  Bi-LSTM (Latest Paper) & 40.9\\
  \hline
  BERT & 52.5\\
  \hline
  DistilBERT & \textbf{62.5}\\
  \hline
\end{tabularx}
\caption{Classification Results for Fully Generated Dataset}
\label{table5}
 \end{center}
\end{table*}

\begin{table*}[ht]
 \begin{center}
 \begin{tabularx}{\textwidth}{|l|X|}
\hline
 Model used & Accuracy \\
 \hline
 Bag of ngrams(1-3), Multinomial Naive Bayes Algorithm

 & 24.2\\
  \hline
 Bag of ngrams(1-3),  Passive Aggressive Classifier Algorithm

 & 30.3 \\ 
  \hline
  Bag of ngrams(1-3), Multinomial Classifier with Hyperparameter (alpha) & 22.7\\
 \hline
 Bag of ngrams(1-3), SVM & 37.9\\
  \hline
  LSTM model & 50.0 \\
  \hline
  Bi-LSTM (Latest Paper) & 47.0\\
  \hline
  BERT & 50.0\\
  \hline
  DistilBERT & \textbf{70.2}\\
  \hline
\end{tabularx}
\caption{Classification Results for Hybrid Dataset}
\label{table6}
 \end{center}
\end{table*}

\begin{table}[!h]
\begin{center}
\begin{tabularx}{\columnwidth}{|X|X|X|X|}

      \hline
      Metric&Full-text Dataset & Hybrid Dataset & Maroniko -lakis et al., 2020 \\
      \hline
      Bi-LSTM & 40.9 & 47.0  & 82.8 \\
      \hline
      BERT  & 52.5 & 50.0 & \textbf{85.7}\\
      \hline
     DistilBERT  & \textbf{62.5} & \textbf{70.2} & 85.5\\
     
      \hline

\end{tabularx}
\caption{Classification Result Comparison for Bi-LSTM, BERT and DistilBERT models}
\label{table7}
 \end{center}
\end{table}



%
%
%
%
%

\section{Conclusion and Future Work}

This paper presents a benchmark proposal for detecting automatically generated research content and describes how it has been produced.
We have considered the detection as a binary classification task and tested various classification algorithms. The results show that the existing state of the art models for classification could provide a maximum accuracy of 70.2\% on our dataset. This result shows that this problem is open and there is room for further improvement in term of accuracy.

Faced with the proliferation of automatically generated content and the risks of scientific misconduct that this represents for the academic world, it is essential to develop efficient detection methods to as to disqualify fake research articles or artificially reviews. The production of datasets of automatically generated academic texts at the level of quality of the best automatic content generation methods (such as deep learning based models) is an indispensable prerequisite. Our proposed benchmark fits into this framework, the next step being the design of classification methods capable of detecting artificially generated academic content.

As a future work, we plan to develop original methods oriented to the task of detecting automatically generated texts and fragments, possibly leveraging knowledge to detect contradictions and out-of-context sentences. Moreover, we hope to increase the size of the dataset by adding more papers to the corpus and thereby re-generating the results.


\section{Acknowledgements}

This work is funded by Université Sorbonne Paris Nord and is partially supported by a public grant overseen by the French National Research Agency (ANR) as part of the program “Investissements d’Avenir” (reference: ANR-10-LABX-0083). 









\section{Bibliographical References}\label{reference}

\bibliographystyle{lrec2022-bib}

\bibliography{lrec2022-MAIN_PAPER}

\begin{thebibliography}{}

\bibitem[\protect\citename{Abd-Elaal \bgroup et al.\egroup
  }2019]{abd2019artificial}
Abd-Elaal, E., Gamage, S., Mills, J., et~al.
\newblock (2019).
\newblock Artificial intelligence is a tool for cheating academic integrity.
\newblock In {\em 30th Annual Conference for the Australasian Association for
  Engineering Education (AAEE 2019): Educators Becoming Agents of Change:
  Innovate, Integrate, Motivate}, page 397. Engineers Australia.

\bibitem[\protect\citename{Adelani \bgroup et al.\egroup
  }2020]{adelani2020generating}
Adelani, D., Mai, H., Fang, F., Nguyen, H., Yamagishi, J., and Echizen, I.
\newblock (2020).
\newblock Generating sentiment-preserving fake online reviews using neural
  language models and their human-and machine-based detection.
\newblock In {\em International Conference on Advanced Information Networking
  and Applications}, pages 1341--1354. Springer.

\bibitem[\protect\citename{Al-Kadhimi and
  L{\"o}wenstr{\"o}m}2020]{al2020identification}
Al-Kadhimi, S. and L{\"o}wenstr{\"o}m, P.
\newblock (2020).
\newblock Identification of machine-generated reviews: 1d cnn applied on the
  gpt-2 neural language model.

\bibitem[\protect\citename{Amancio}2015]{amancio2015comparing}
Amancio, D.~R.
\newblock (2015).
\newblock Comparing the topological properties of real and artificially
  generated scientific manuscripts.
\newblock {\em Scientometrics}, 105(3):1763--1779.

\bibitem[\protect\citename{Bakhtin \bgroup et al.\egroup
  }2019]{bakhtin2019real}
Bakhtin, A., Gross, S., Ott, M., Deng, Y., Ranzato, M., and Szlam, A.
\newblock (2019).
\newblock Real or fake? learning to discriminate machine from human generated
  text.
\newblock {\em arXiv preprint arXiv:1906.03351}.

\bibitem[\protect\citename{Bhat and Parthasarathy}2020]{bhat2020effectively}
Bhat, M.~M. and Parthasarathy, S.
\newblock (2020).
\newblock How effectively can machines defend against machine-generated fake
  news? an empirical study.
\newblock In {\em Proceedings of the First Workshop on Insights from Negative
  Results in NLP}, pages 48--53.

\bibitem[\protect\citename{Bojanowski \bgroup et al.\egroup
  }2017]{bojanowski2017enriching}
Bojanowski, P., Grave, E., Joulin, A., and Mikolov, T.
\newblock (2017).
\newblock Enriching word vectors with subword information.
\newblock {\em Transactions of the Association for Computational Linguistics},
  5:135--146.

\bibitem[\protect\citename{Cabanac and Labb{\'e}}2021]{cabanac2021prevalence}
Cabanac, G. and Labb{\'e}, C.
\newblock (2021).
\newblock Prevalence of nonsensical algorithmically generated papers in the
  scientific literature.
\newblock {\em Journal of the Association for Information Science and
  Technology}.

\bibitem[\protect\citename{Clark \bgroup et al.\egroup }2021]{clark2021all}
Clark, E., August, T., Serrano, S., Haduong, N., Gururangan, S., and Smith,
  N.~A.
\newblock (2021).
\newblock All that's' human'is not gold: Evaluating human evaluation of
  generated text.
\newblock {\em arXiv preprint arXiv:2107.00061}.

\bibitem[\protect\citename{Cortes and Vapnik}1995]{cortes1995support}
Cortes, C. and Vapnik, V.
\newblock (1995).
\newblock Support-vector networks.
\newblock {\em Machine learning}, 20(3):273--297.

\bibitem[\protect\citename{Dai \bgroup et al.\egroup
  }2019]{Dai-Yang-Yang-Carbonell-Le-Salakhutdinov-19}
Dai, Z., Yang, Z., Yang, Y., Carbonell, J., Le, Q., and Salakhutdinov, R.
\newblock (2019).
\newblock Transformer-xl: Attentive language models beyond a fixed-length
  context.

\bibitem[\protect\citename{Devlin \bgroup et al.\egroup
  }2018]{Devlin-Chang-Lee-Toutanova-18}
Devlin, J., Chang, M., Lee, K., and Toutanova, K.
\newblock (2018).
\newblock Bert: Pre-training of deep bidirectional transformers for language
  understanding.

\bibitem[\protect\citename{Dugan \bgroup et al.\egroup }2020]{dugan2020roft}
Dugan, L., Ippolito, D., Kirubarajan, A., and Callison-Burch, C.
\newblock (2020).
\newblock Roft: A tool for evaluating human detection of machine-generated
  text.
\newblock {\em arXiv preprint arXiv:2010.03070}.

\bibitem[\protect\citename{Fagni \bgroup et al.\egroup
  }2021]{fagni2021tweepfake}
Fagni, T., Falchi, F., Gambini, M., Martella, A., and Tesconi, M.
\newblock (2021).
\newblock Tweepfake: About detecting deepfake tweets.
\newblock {\em Plos one}, 16(5):e0251415.

\bibitem[\protect\citename{Faris \bgroup et al.\egroup
  }2017]{faris2017partisanship}
Faris, R., Roberts, H., Etling, B., Bourassa, N., Zuckerman, E., and Benkler,
  Y.
\newblock (2017).
\newblock Partisanship, propaganda, and disinformation: Online media and the
  2016 us presidential election.
\newblock {\em Berkman Klein Center Research Publication}, 6.

\bibitem[\protect\citename{Gehrmann \bgroup et al.\egroup
  }2019]{gehrmann2019gltr}
Gehrmann, S., Strobelt, H., and Rush, A.
\newblock (2019).
\newblock Gltr: Statistical detection and visualization of generated text.
\newblock {\em arXiv preprint arXiv:1906.04043}.

\bibitem[\protect\citename{Harada \bgroup et al.\egroup
  }2021]{harada2021discrimination}
Harada, A., Bollegala, D., and Chandrasiri, N.~P.
\newblock (2021).
\newblock Discrimination of human-written and human and machine written
  sentences using text consistency.
\newblock In {\em 2021 International Conference on Computing, Communication,
  and Intelligent Systems (ICCCIS)}, pages 41--47. IEEE.

\bibitem[\protect\citename{Harris}1954]{harris1954distributional}
Harris, Z.~S.
\newblock (1954).
\newblock Distributional structure.
\newblock {\em Word}, 10(2-3):146--162.

\bibitem[\protect\citename{Ippolito \bgroup et al.\egroup
  }2019a]{ippolito2019automatic}
Ippolito, D., Duckworth, D., Callison-Burch, C., and Eck, D.
\newblock (2019a).
\newblock Automatic detection of generated text is easiest when humans are
  fooled.
\newblock {\em arXiv preprint arXiv:1911.00650}.

\bibitem[\protect\citename{Ippolito \bgroup et al.\egroup
  }2019b]{ippolito2019human}
Ippolito, D., Duckworth, D., Callison-Burch, C., and Eck, D.
\newblock (2019b).
\newblock Human and automatic detection of generated text.

\bibitem[\protect\citename{Jawahar}]{jawahardetecting}
Jawahar, G.
\newblock ().
\newblock Detecting human written text from machine generated text by modeling
  discourse coherence.

\bibitem[\protect\citename{Jawahar \bgroup et al.\egroup
  }2020]{jawahar2020automatic}
Jawahar, G., Abdul-Mageed, M., and Lakshmanan, L.~V.
\newblock (2020).
\newblock Automatic detection of machine generated text: A critical survey.
\newblock {\em arXiv preprint arXiv:2011.01314}.

\bibitem[\protect\citename{Lample and Conneau}2019]{lample2019cross}
Lample, G. and Conneau, A.
\newblock (2019).
\newblock Cross-lingual language model pretraining.

\bibitem[\protect\citename{Lavoie and
  Krishnamoorthy}2010]{lavoie2010algorithmic}
Lavoie, A. and Krishnamoorthy, M.
\newblock (2010).
\newblock Algorithmic detection of computer generated text.
\newblock {\em arXiv preprint arXiv:1008.0706}.

\bibitem[\protect\citename{Lewis \bgroup et al.\egroup }2019]{lewis2019bart}
Lewis, M., Liu, Y., Goyal, N., Ghazvininejad, M., Mohamed, A., Levy, O.,
  Stoyanov, V., and Zettlemoyer, L.
\newblock (2019).
\newblock Bart: Denoising sequence-to-sequence pre-training for natural
  language generation, translation, and comprehension.

\bibitem[\protect\citename{Lin}2004]{lin2004rouge}
Lin, C.-Y.
\newblock (2004).
\newblock Rouge: A package for automatic evaluation of summaries.
\newblock In {\em Text summarization branches out}, pages 74--81.

\bibitem[\protect\citename{Liu \bgroup et al.\egroup }2019]{liu2019roberta}
Liu, Y., Ott, M., Goyal, N., Du, J., Joshi, M., Chen, D., Levy, O., Lewis, M.,
  Zettlemoyer, L., and Stoyanov, V.
\newblock (2019).
\newblock Roberta: A robustly optimized bert pretraining approach.
\newblock {\em arXiv preprint arXiv:1907.11692}.

\bibitem[\protect\citename{Maronikolakis \bgroup et al.\egroup
  }2020]{maronikolakis2020identifying}
Maronikolakis, A., Schutze, H., and Stevenson, M.
\newblock (2020).
\newblock Identifying automatically generated headlines using transformers.
\newblock {\em arXiv preprint arXiv:2009.13375}.

\bibitem[\protect\citename{Nguyen and Labb{\'e}}2016]{nguyen2016engineering}
Nguyen, M.~T. and Labb{\'e}, C.
\newblock (2016).
\newblock Engineering a tool to detect automatically generated papers.
\newblock In {\em BIR 2016 Bibliometric-enhanced Information Retrieval}.

\bibitem[\protect\citename{Papineni \bgroup et al.\egroup
  }2002]{papineni2002bleu}
Papineni, K., Roukos, S., Ward, T., and Zhu, W.-J.
\newblock (2002).
\newblock Bleu: a method for automatic evaluation of machine translation.
\newblock In {\em Proceedings of the 40th annual meeting of the Association for
  Computational Linguistics}, pages 311--318.

\bibitem[\protect\citename{P{\'e}rez-Rosas \bgroup et al.\egroup
  }2017]{perez2017automatic}
P{\'e}rez-Rosas, V., Kleinberg, B., Lefevre, A., and Mihalcea, R.
\newblock (2017).
\newblock Automatic detection of fake news.
\newblock {\em arXiv preprint arXiv:1708.07104}.

\bibitem[\protect\citename{Radford \bgroup et al.\egroup
  }2018]{Radford-Narasimhan-Salimans-Sutskever-18}
Radford, A., Narasimhan, K., Salimans, T., and Sutskever, I.
\newblock (2018).
\newblock Improving language understanding by generative pre-training.

\bibitem[\protect\citename{Radford \bgroup et al.\egroup
  }2019]{radford2019language}
Radford, A., Wu, J., Child, R., Luan, D., Amodei, D., and Sutskever, I.
\newblock (2019).
\newblock Language models are unsupervised multitask learners.
\newblock volume~1, page~9.

\bibitem[\protect\citename{Sanh \bgroup et al.\egroup
  }2019]{sanh2019distilbert}
Sanh, V., Debut, L., Chaumond, J., and Wolf, T.
\newblock (2019).
\newblock Distilbert, a distilled version of bert: smaller, faster, cheaper and
  lighter.
\newblock {\em arXiv preprint arXiv:1910.01108}.

\bibitem[\protect\citename{Solaiman \bgroup et al.\egroup
  }2019]{solaiman2019release}
Solaiman, I., Brundage, M., Clark, J., Askell, A., Herbert-Voss, A., Wu, J.,
  Radford, A., Krueger, G., Kim, J., Kreps, S., et~al.
\newblock (2019).
\newblock Release strategies and the social impacts of language models.
\newblock {\em arXiv preprint arXiv:1908.09203}.

\bibitem[\protect\citename{Uchendu \bgroup et al.\egroup
  }2020]{uchendu2020authorship}
Uchendu, A., Le, T., Shu, K., and Lee, D.
\newblock (2020).
\newblock Authorship attribution for neural text generation.
\newblock In {\em Conf. on Empirical Methods in Natural Language Processing
  (EMNLP)}.

\bibitem[\protect\citename{Varshney \bgroup et al.\egroup
  }2020]{varshney2020limits}
Varshney, L.~R., Keskar, N.~S., and Socher, R.
\newblock (2020).
\newblock Limits of detecting text generated by large-scale language models.
\newblock In {\em 2020 Information Theory and Applications Workshop (ITA)},
  pages 1--5. IEEE.

\bibitem[\protect\citename{Vaswani \bgroup et al.\egroup
  }2017]{Vaswani-Shazeer-Parmar-Uszkoreit-Jones-Gomez-Kaiser-Polosukhin-17}
Vaswani, A., Shazeer, N., Parmar, N., Uszkoreit, J., Jones, L., Gomez, A.~N.,
  Kaiser, {\L}., and Polosukhin, I.
\newblock (2017).
\newblock Attention is all you need.
\newblock In {\em Advances in neural information processing systems}, pages
  5998--6008.

\bibitem[\protect\citename{Vijayaraghavan \bgroup et al.\egroup
  }2020]{vijayaraghavan2020fake}
Vijayaraghavan, S., Wang, Y., Guo, Z., Voong, J., Xu, W., Nasseri, A., Cai, J.,
  Li, L., Vuong, K., and Wadhwa, E.
\newblock (2020).
\newblock Fake news detection with different models.
\newblock {\em arXiv preprint arXiv:2003.04978}.

\bibitem[\protect\citename{Vosoughi \bgroup et al.\egroup
  }2018]{vosoughi2018spread}
Vosoughi, S., Roy, D., and Aral, S.
\newblock (2018).
\newblock The spread of true and false news online.
\newblock {\em Science}, 359(6380):1146--1151.

\bibitem[\protect\citename{Wang \bgroup et al.\egroup
  }2019]{wang2019paperrobot}
Wang, Q., Huang, L., Jiang, Z., Knight, K., Ji, H., Bansal, M., and Luan, Y.
\newblock (2019).
\newblock Paperrobot: Incremental draft generation of scientific ideas.
\newblock {\em arXiv preprint arXiv:1905.07870}.

\bibitem[\protect\citename{Wang \bgroup et al.\egroup
  }2020]{wang2020reviewrobot}
Wang, Q., Zeng, Q., Huang, L., Knight, K., Ji, H., and Rajani, N.
\newblock (2020).
\newblock Reviewrobot: Explainable paper review generation based on knowledge
  synthesis.
\newblock {\em arXiv preprint arXiv:2010.06119}.

\bibitem[\protect\citename{Wardle and Derakhshan}2017]{wardle2017information}
Wardle, C. and Derakhshan, H.
\newblock (2017).
\newblock Information disorder: Toward an interdisciplinary framework for
  research and policy making.
\newblock {\em Council of Europe}, 27.

\bibitem[\protect\citename{Williams and Giles}2015]{williams2015use}
Williams, K. and Giles, C.~L.
\newblock (2015).
\newblock On the use of similarity search to detect fake scientific papers.
\newblock In {\em International Conference on Similarity Search and
  Applications}, pages 332--338. Springer.

\bibitem[\protect\citename{Wolf \bgroup et al.\egroup
  }2020]{wolf-etal-2020-transformers}
Wolf, T., Debut, L., Sanh, V., Chaumond, J., Delangue, C., Moi, A., Cistac, P.,
  Rault, T., Louf, R., Funtowicz, M., Davison, J., Shleifer, S., von Platen,
  P., Ma, C., Jernite, Y., Plu, J., Xu, C., Scao, T.~L., Gugger, S., Drame, M.,
  Lhoest, Q., and Rush, A.~M.
\newblock (2020).
\newblock Transformers: State-of-the-art natural language processing.
\newblock In {\em Proceedings of the 2020 Conference on Empirical Methods in
  Natural Language Processing: System Demonstrations}, pages 38--45, Online,
  October. Association for Computational Linguistics.

\bibitem[\protect\citename{Wolff and Wolff}2020]{wolff2020attacking}
Wolff, M. and Wolff, S.
\newblock (2020).
\newblock Attacking neural text detectors.
\newblock {\em arXiv preprint arXiv:2002.11768}.

\bibitem[\protect\citename{Xiong and Huang}2009]{xiong2009effective}
Xiong, J. and Huang, T.
\newblock (2009).
\newblock An effective method to identify machine automatically generated
  paper.
\newblock In {\em 2009 Pacific-Asia Conference on Knowledge Engineering and
  Software Engineering}, pages 101--102. IEEE.

\bibitem[\protect\citename{Yang \bgroup et al.\egroup }2019]{yang2019xlnet}
Yang, Z., Dai, Z., Yang, Y., Carbonell, J., Salakhutdinov, R., and Le, Q.~V.
\newblock (2019).
\newblock Xlnet: Generalized autoregressive pretraining for language
  understanding.
\newblock volume~32.

\bibitem[\protect\citename{Zellers \bgroup et al.\egroup
  }2019]{zellers2019defending}
Zellers, R., Holtzman, A., Rashkin, H., Bisk, Y., Farhadi, A., Roesner, F., and
  Choi, Y.
\newblock (2019).
\newblock Defending against neural fake news.
\newblock {\em arXiv preprint arXiv:1905.12616}.

\end{thebibliography}


\end{document}